# Dialogue Act Patterns in GenAI-Mediated L2 Oral Practice: A Sequential Analysis of Learner–Chatbot Interactions

Liqun He[1], Shijun (Cindy) Chen[2], Mutlu Cukurova[1], Manolis Mavrikis[1]

[1]University College London, London, United Kingdom
`{liqun.he,m.cukurova,m.mavrikis}@ucl.ac.uk`
[2]University of Hong Kong, Hong Kong, China
`shijunchen@connect.hku.hk`

**Abstract.** While generative AI (GenAI) voice chatbots offer scalable opportunities for second language (L2) oral practice, the interactional processes related to learners' gains remain underexplored. This study investigates dialogue act (DA) patterns in interactions between Grade 9 Chinese English as a foreign language (EFL) learners and a GenAI voice chatbot over a 10-week intervention. Seventy sessions from 12 students were annotated by human coders using a pedagogy-informed coding scheme, yielding 6,957 coded DAs. DA distributions and sequential patterns were compared between high- and low-progress sessions. At the DA level, high-progress sessions showed more learner-initiated questions, whereas low-progress sessions exhibited higher rates of clarification-seeking, indicating greater comprehension difficulty. At the sequential level, high-progress sessions were characterised by more frequent prompting-based corrective feedback sequences, consistently positioned after learner responses, highlighting the role of feedback type and timing in effective interaction. Overall, these findings underscore the value of a dialogic lens in GenAI chatbot design, contribute a pedagogy-informed DA coding framework, and inform the design of adaptive GenAI chatbots for L2 education.

## 1 Introduction

The increasing integration of generative AI (GenAI) into language education has created new opportunities for second language (L2) oral practice. By generating human-like responses without predefined templates, these systems enable immersive conversational practice unrestricted by time or location, addressing the challenge of limited speaking opportunities that many EFL learners face where English is not widely used outside the classroom [1, 2].

However, knowing that GenAI-based chatbots have the potential to improve oral proficiency does not explain the mechanism underlying these outcomes. While systematic reviews and meta-analyses have demonstrated positive effects of GenAI voice chatbots on various dimensions of oral proficiency, including fluency, complexity, and pronunciation [1–3], few studies have examined the *dialogic processes* underlying these outcomes. Given the central role of communication in developing oral proficiency [4, 5], focusing solely on pre–post gains risks overlooking the interactional mechanisms through which chatbots facilitate, or potentially hinder, L2 oral development.

Thus, *a dialogic lens* is essential for a fine-grained understanding of the interactional mechanisms through which learning is mediated. Based on Speech Act Theory, this study adopts dialogue analysis to analyse both the dialogue acts (DAs) and their sequence associated with learning gains in the context of GenAI-mediated oral practice. A pedagogy-informed DA coding scheme is developed and introduced first. Then, learner-GenAI interaction sessions were annotated at the turn level with DA labels by two human coders to capture pedagogical and interactional intentions. After grouping all sessions based on learners' pre-post oral proficiency gains, descriptive analyses and statistical comparisons were used to examine DA-level patterns associated with learning outcomes. Sequential pattern mining is then applied to identify predominant DA patterns and to examine how these patterns are associated with differential learning outcomes. Specifically, the study addresses the following two research questions:

- RQ1: What differences in DA frequencies are associated with high vs. low learning gains?
- RQ2: What sequential DA patterns are associated with high vs. low learning gains?

This study contributes to AIED research in three ways. First, by integrating language teaching theories with dialogue act analysis, it proposes a systematic DA coding scheme for examining dialogue process for GenAI-mediated language learning studies. Second, it offers exploratory evidence that certain DA frequencies and sequential patterns are associated with higher versus lower oral proficiency gains, highlighting the value of adopting a dialogic lens in educational chatbot research. Third, these findings offer design-relevant insights that can inform the development of more adaptive voice-based tutoring systems in future work.

## 2      Theoretical Basis and Related Work

### 2.1     Speech Act Theory and the Dialogic Lens

This study draws on Speech Act Theory as the theoretical foundation for analysing pedagogical processes in GenAI–learner interaction. Speech Act Theory posits that every utterance performs an action by conveying not only expressed content but also underlying intentions, and these intentions can be inferred from the utterance's form and its context [6]. These intentions are formalised as dialogue acts (DAs), which capture the functional role of an utterance rather than its surface semantic meaning [6, 7]. Identifying DAs therefore requires moving beyond literal meaning to examine what speakers are attempting to accomplish [7]. For example, an utterance such as "I am hungry" does not merely describe a physical state but may implicitly function as a request, a signal of desire to eat, or an invitation to change activities.

From such a dialogic perspective, pedagogical purpose can be inferred from observable dialogic behaviours, as represented by DAs. These pedagogical intentions are conveyed through what teachers say and when they say it [8]. Through different forms of dialogue, instructors pursue a range of pedagogical goals, such as providing hints, scaffolding, and checking understanding [9]. Conceptualising different forms of utterances as different DAs therefore provides a theoretical link between pedagogical intentions and observable dialogic behaviours, enabling the systematic analysis of educational dialogue. In the L2 learning domain, speech act analysis has been widely used to examine interactional processes in various contexts [10]. Importantly, pedagogical strategies are not always realised through isolated DAs but often unfold across DA *sequences*. For example, scaffolding becomes visible only through a sequence of prompts and hints that progressively shift responsibility to the learner [11]. Thus, analysing DAs and their sequences enables the reconstruction of higher-order constructs that cannot be captured at the single-turn level [12].

In GenAI-mediated oral practice, analysing DAs and their sequences enables examination of how pedagogical intentions are realised and how learning is mediated by these interactions. On this basis, Speech Act Theory provides the theoretical grounding for the dialogic lens adopted in this study, guiding the subsequent analysis.

### 2.2   Related Work

**Dialogue Analysis in Human Tutoring.** Before the emergence of LLM-based educational systems, dialogue analysis had been widely used to examine pedagogical processes in human tutoring, yielding robust evidence on how interactional patterns relate to learning outcomes. For example, [13] analysed 50 one-to-one tutoring sessions in mathematics, chemistry, and physics using BERT-based DA annotation combined with sequential pattern mining. They found that more successful sessions featured a higher proportion of thought-provoking questions and informational hints, alongside fewer irrelevant tutor moves. Similarly, [14] examined over 2,000 minutes of online mathematics tutoring using manual annotations of dialogic behaviours. Their sequential analyses identified effective strategies such as prompting learner self-correction and allowing pauses for reflection, and these patterns reliably distinguished more effective tutors from less effective ones.

These studies demonstrate the value of dialogue analysis in human tutoring. However, most prior work has focused on STEM domains, with very limited use of dialogic analysis in language tutoring. This gap is significant because language learning is inherently interaction-driven [15]. Further, as AI-generated language differs systematically from human-authored language, it remains unclear whether pedagogical patterns identified in human contexts apply to GenAI-mediated contexts, highlighting the need for more research in this emerging learning setting [16].

**Dialogue Analysis in GenAI-Mediated Learning.** With the increasing use of GenAI in education, a growing body of research has begun to apply dialogue analysis to examine pedagogical processes in AI-mediated learning environments. For example, [17] analysed learners' interactions with a GenAI-based Socratic agent by coding reflective thinking in dialogue turns and modelling how these categories co-occur across conversations. Their results showed that different agent designs were associated with different patterns of reflective engagement. In language education, [18] examined L2 children's interactions with a GenAI agent during dialogic reading activities, analysing question–response–feedback structures to illustrate how instructional dialogue design influences learner engagement. More broadly, emerging work in L2 learning suggests that learning outcomes in GenAI-mediated settings are influenced not only by linguistic input, but also by interactional features of the dialogue itself [19].

While these studies demonstrate the value of dialogue analysis for understanding learning processes, several limitations remain. First, dialogue is often analysed at a relatively coarse level, with limited attention to fine-grained, turn-level analysis, which restricts detailed accounts of how learning is mediated through interaction. Second, relatively few studies adopt pedagogy-informed coding frameworks that capture the interactional intentions of both learners and AI interlocutors, limiting systematic understanding of the dialogue process. Finally, although learning outcomes are commonly reported, explicit analyses linking sequential DA patterns to differential learning gains remain limited, constraining the use of interactional evidence to refine chatbot design.

## 3   Method

### 3.1   Dataset

Ethical approval was obtained from the first author's institutional ethics committee. This study draws on a de-identified dataset from a larger intervention that examines the effects of GenAI voice chatbots on L2 oral proficiency development. The dataset comprised 70 GenAI-mediated oral practice sessions collected from 12 Grade 9 students (aged 14–15) at an international school in central China. All participants were native Mandarin speakers with English proficiency ranging from A2 to B1 according to the CEFR (Common European Framework of Reference for Languages).

The GenAI voice chatbot was purpose-built by [20] to support L2 oral development through spoken interaction and corrective feedback (CF). Its prompts instructed the system to maintain conversational engagement while providing feedback on learners' linguistic errors (e.g., grammar, lexical choice). Students completed weekly 35-minute practice sessions in the school IT lab using individual computers and noise-cancelling headphones. Conversations were open-ended, with topics rotating across daily life, travel, food, and study.

Learners' oral proficiency was assessed before and after the intervention using TOEFL Practice Online (TPO) independent speaking tasks. Pre- and post-tests were matched for task type and difficulty but differed in content to minimise test–retest effects. All learner–chatbot interactions were audio-recorded, transcribed, and stored in a secure cloud database. Personal identifiers were removed through pseudonymisation (e.g., [learner_01]), and no identifiable information was retained.

Given that this study aims to examine DAs and their patterns associated with learners' oral proficiency gains, the dataset was then processed in three steps:

**Step 1: Oral Proficiency Scoring and Gain Calculation.** Learners' pre- and post-test performances were evaluated by trained human raters using the Complexity–Accuracy–Fluency (CAF) framework [21]. A combined measure was computed based on six indicators: lexical complexity (non-A1 words per 100 words), grammatical complexity (clauses per AS-unit), lexical accuracy (lexical misuses per 100 words), grammatical accuracy (grammatical errors per 100 words), speed fluency (words per articulation time), and breakdown and repair fluency (frequency and duration of pauses, repetitions, and self-corrections). For indicators requiring annotation, two experienced English teachers independently coded the transcripts. Interrater reliability was high across measures, with ICC values ranging from 0.78 to 0.84. All indicators were z-score standardised and averaged to obtain overall oral proficiency scores at pre-test and post-test, with learning gains calculated as post–pre differences.

**Step 2: Dialogue data preprocessing**. All anonymised dialogue data were cleaned by removing empty turns due to recording failures. The final dataset comprised 70 valid sessions and 4,922 dialogue turns ($M$ = 70.31 turns per session), yielding a dense, fine-grained corpus.

**Step 3: Session-level Grouping by Learning Gains**. All sessions were grouped according to the pre–post oral proficiency gains of the learners who produced them. Sessions from learners with larger gains were classified as high-progress (HP) (35 sessions from 6 learners), while sessions from learners with smaller gains were classified as low-progress (LP) (35 sessions from 6 learners).

### 3.2   Dialogue Act Coding Scheme and Human Annotation

A theory-informed, deductive DA coding scheme was developed to systematically capture both meaning-focused and form-focused dimensions of dialogic interaction. The scheme is grounded in Communicative Language Teaching (CLT) and Focus on Form (FonF). CLT emphasises meaning-focused communication as a central approach for language learning [4]. Building on CLT, FonF highlights the pedagogical value of brief, timely attention to linguistic form within the communicative activity, enabling learners to notice and correct errors without disrupting meaning-focused interaction [22].

Specifically, meaning-focused DAs were organised around three pedagogical functions: inviting, sustaining, and content feedback. Inviting DAs capture interactional moves that open opportunities for meaningful communication by eliciting new information rather than checking comprehension. These include referential questions (Q) and interaction-opening moves such as greetings (G) and topic shifts (T) [5, 23]. Sustaining DAs capture behaviours that support learners in expressing intended meaning and maintaining interactional flow, including seeking clarification (S) and expressing agreement (A) or disagreement (D) [5, 24]. Instances of misinterpretation without clarification-seeking (M) were also coded, as they are particularly salient in GenAI-mediated oral interaction. Content feedback DAs capture responses to meaning rather than form, primarily through responding (R) to expressed ideas by answering, elaborating on, or extending meaning. In CLT, such feedback functions to sustain discourse and elicit further information rather than to correct linguistic errors [5, 23].

Form-focused DAs capture CF, defined as interactional moves that draw learners' attention to linguistic form within ongoing communication [25, 26]. These include recasts (Cr), which implicitly reformulate erroneous utterances; prompts (Cp), which signal errors and encourage self-correction without explicit explanation; and explicit correction (Ce), which directly identifies errors and provides the target form [25, 26].

Grounded in these pedagogical principles, the resulting coding scheme (Table 1) captures interactional moves theoretically central to L2 oral development in GenAI-mediated oral practice.

**Table 1.** Coding Scheme for Identifying DAs in GenAI-Mediated Oral Practice

| Dimension | Linguistic Function | Code | Dialogue Act | Description |
|---|---|---|---|---|
| Meaning-focused | Inviting | Q | Referential Question | Questions eliciting new information, elaboration, or personal experience rather than checking comprehension. |
| | | G | Greeting/ Closing | Utterances initiating or concluding the dialogue (e.g., greetings, farewells). |
| | | T | Topic Shifting | Initiating a shift from the current topic to a new topic to re-invite engagement. |
| | Sustaining | S | Seeking Clarification | Requesting clarification of the interlocutor's previous utterance to negotiate meaning. |
| | | A | Agreement | Expressing empathy or agreement with the interlocutor's viewpoint. |
| | | D | Disagreement | Expressing disagreement with the interlocutor's viewpoint in an authentic manner. |
| | | M | Misinterpretation | Responses reflecting misunderstanding of the interlocutor's intended meaning. |
| | Content Feedback | R | Response | Answers or elaborations responding to previous utterances. |
| Form-focused | Corrective Feedback | Cr | Recast | Implicit reformulation of an erroneous utterance using the correct linguistic form. |
| | | Cp | Prompting | Signalling an error and encouraging self-correction without explicit explanation. |
| | | Ce | Explicit Correction | Explicitly pointing out the error, explaining it, and providing the correct form. |

**Human Annotation Procedure.** Human annotation was conducted to identify DA tags. Coding was performed at the turn level, with each speaking turn treated as the unit of analysis. Turn-level coding is widely used in dialogic research, as speaker changes can be reliably identified and support high inter-coder reliability [27]. Because a single turn may serve multiple interactional functions, up to two DA codes were assigned per turn. Prior to formal coding, the first author conducted a two-hour training session with both coders to introduce the coding scheme and jointly annotate the first 200 turns, ensuring a shared understanding of each code. The remaining turns were then independently annotated by two expert EFL teachers (L1 Mandarin).

Following annotation, DA labels were prefixed by speaker role to enable role-specific analyses, with [s] indicating the student and [t] indicating the chatbot (e.g., [s]R denotes a student response).

**Inter-coder Reliability.** Inter-coder reliability was assessed using a multi-label procedure, in which each DA code was treated as a binary variable indicating its presence or absence within a turn. This approach is appropriate for dialogue coding schemes that allow multiple codes per utterance [28]. Agreement was estimated using Cohen's kappa, calculated separately for each code. Kappa values ranged from fair to almost perfect ($\kappa$ = .25–.88). Three codes fell below the .60 threshold: Ce ($\kappa$ = .50), M ($\kappa$ = .45), and D ($\kappa$ = .25). These agreement variations were mainly attributable to three sources of ambiguity: (1) identifying different CF types, (2) distinguishing learner misinterpretation (M) from minimally engaged responses (R), and (3) differentiating disagreement (D) from negative responses (R). The first author reviewed all discrepant turns and made final coding decisions in accordance with the coding scheme. Any remaining uncertainties were resolved through discussion with the other two coders.

### 3.3  Data Analysis

**DA Comparison (RQ1).** DA comparison was conducted at the individual DA level to examine differences in DA frequencies between the HP and LP. Role-specific DA frequencies were compared using chi-square tests, with Holm–Bonferroni correction applied to control for multiple comparisons.

**Sequential Pattern Comparison (RQ2).** To address RQ2, sequential pattern analysis was conducted in three steps:

*Step 1.* Frequent DA sequences were mined using the CM-SPAM algorithm [29], with a minimum pattern length of 2, a maximum length of 4, a maximum gap of 1, and a minimum support threshold of 20%. Here, support refers to the number of sessions in which a given DA sequence occurs. The maximum pattern length was set to capture complete interactional exchanges while avoiding overly long and sparse sequences. This step yielded 107 frequent DA patterns.

*Step 2.* To focus on patterns showing meaningful group-level contrasts, the mined sequences were filtered using a support difference threshold of 10 between the HP and LP groups. This threshold was used to retain patterns with meaningful prevalence differences while reducing noise and redundancy commonly produced by frequent sequence mining [30]. This step resulted in 9 DA patterns.

*Step 3.* Group differences in pattern prevalence were tested using permutation tests with Holm–Bonferroni correction. This non-parametric approach was adopted because multiple sessions were nested within individual learners, violating the assumption of independent observations. Student-level shuffling was therefore used to preserve within-learner dependency structures. Permutation tests can provide exact control of false positives and rely only on the assumption of exchangeability under the null hypothesis, and are well suited to small sample sizes or non-normal data distributions [30, 31].

## 4      Results

### 4.1      DA Frequencies Associated with Learning Gains (RQ1)

Table 2 presents overall DA distributions and compares DA frequencies between the HP and LP groups. Across both groups, the three most frequent codes were identified: chatbot questions ([t]Q, 31.4%), learner responses ([s]R, 27.2%), and chatbot agreement ([t]A, 19.0%), together accounting for 77.6% of all coded DAs.

Chi-square tests with Holm–Bonferroni correction further revealed two significant group differences. First, HP learners asked more questions ([s]Q) than LP learners (HP: 2.1% vs. LP: 1.0%), $\chi^2(1) = 11.37$, $p = .001$, adjusted $p = .013$. In contrast, LP learners exhibited significantly higher frequencies of clarification-seeking ([s]S) than HP learners (LP: 2.5% vs. HP: 1.3%), $\chi^2(1) = 13.78$, $p < .001$, adjusted $p = .004$. No other DA codes showed significant group differences after correction.

Table 2. DA Distributions Overall and by Group (HP vs. LP), Sorted by Total Frequency

| DA (with roles) | Overall | | HP | | LP | | $\chi^2(1)$ | p | p_adj |
|---|---|---|---|---|---|---|---|---|---|
| | n | % | n | % | n | % | | | |
| [t]Q | 2186 | 31.4 | 1135 | 31.5 | 1051 | 31.3 | 0.02 | 0.902 | 1.000 |
| [s]R | 1895 | 27.2 | 985 | 27.3 | 910 | 27.1 | 0.03 | 0.868 | 1.000 |
| [t]A | 1323 | 19.0 | 662 | 18.4 | 661 | 19.7 | 1.92 | 0.166 | 1.000 |
| [t]S | 331 | 4.8 | 163 | 4.5 | 168 | 5.0 | 0.8 | 0.372 | 1.000 |
| [t]R | 179 | 2.6 | 97 | 2.7 | 82 | 2.4 | 0.33 | 0.565 | 1.000 |
| [t]Cp | 169 | 2.4 | 98 | 2.7 | 71 | 2.1 | 2.42 | 0.12 | 1.000 |
| **[s]S** | **128** | **1.8** | **45** | **1.3** | **83** | **2.5** | **13.78** | **0** | **.004*** |
| **[s]Q** | **110** | **1.6** | **75** | **2.1** | **35** | **1.0** | **11.37** | **0.001** | **.013*** |
| [s]M | 96 | 1.4 | 49 | 1.4 | 47 | 1.4 | 0 | 0.964 | 1.000 |
| [s]G | 93 | 1.3 | 40 | 1.1 | 53 | 1.6 | 2.56 | 0.109 | 1.000 |
| [t]G | 82 | 1.2 | 35 | 1.0 | 47 | 1.4 | 2.4 | 0.121 | 1.000 |
| [t]Ce | 78 | 1.1 | 48 | 1.3 | 30 | 0.9 | 2.62 | 0.106 | 1.000 |
| [s]T | 68 | 1.0 | 36 | 1.0 | 32 | 1.0 | 0 | 0.945 | 1.000 |
| [t]Cr | 66 | 0.9 | 39 | 1.1 | 27 | 0.8 | 1.14 | 0.285 | 1.000 |

| | | | | | | | | |
|---|---|---|---|---|---|---|---|---|
| [t]T | 54 | 0.8 | 28 | 0.8 | 26 | 0.8 | 0 | 1.000 | 1.000 |
| [s]D | 47 | 0.7 | 31 | 0.9 | 16 | 0.5 | 3.25 | 0.071 | 1.000 |
| [s]A | 19 | 0.3 | 11 | 0.3 | 8 | 0.2 | 0.09 | 0.762 | 1.000 |
| [t]D | 17 | 0.2 | 13 | 0.4 | 4 | 0.1 | 3.23 | 0.073 | 1.000 |
| [t]M | 16 | 0.2 | 13 | 0.4 | 3 | 0.1 | 4.45 | 0.035 | 0.592 |
| SUM | 6957 | 100 | 3603 | 100 | 3354 | 100 | | | |

*Note.* *$p < .05$ (Holm–Bonferroni–corrected)

## 4.2   Sequential DA Patterns Associated with Learning Gains (RQ2)

Table 3 presents patterns showing statistically significant or marginally significant differences between groups after Holm–Bonferroni correction. Two patterns were statistically significant, and four showed marginal significance ($.05 <$ adjusted $p < .10$). Marginally significant patterns are reported given the exploratory nature of the study.

**Table 3.** Group Differences in Sequential DA Patterns (HP vs. LP)

| # | Patterns | HP SUP | LP SUP | SUP DIFF | p | p_adj |
|---|---|---|---|---|---|---|
| 1 | [t]Q → [s]R → [t]Cp | 28 | 15 | 13 | .005 | .041* |
| 2 | [t]Q → [s]R → [t]Cp → [t]Q | 28 | 15 | 13 | .005 | .041* |
| 3 | [s]R → [t]Cp | 28 | 17 | 11 | .009 | .063† |
| 4 | [s]R → [t]Cp → [t]Q | 28 | 17 | 11 | .009 | .063† |
| 5 | [s]R → [t]Cp → [t]Q → [s]R | 26 | 15 | 11 | .009 | .063† |
| 6 | [t]Q → [s]D | 16 | 5 | 11 | .020 | .079† |
| 7 | [t]A → [t]Q → [s]R → [t]Cp | 24 | 14 | 10 | .035 | .104 |
| 8 | [t]R → [t]Q → [s]R → [t]A | 11 | 21 | -10 | .041 | .104 |
| 9 | [t]Cp → [t]Q → [s]R → [t]A | 23 | 13 | 10 | .053 | .104 |

*Note.* SUP DIFF = difference in pattern support between groups (HP − LP).
*$p < .05$ (Holm–Bonferroni–corrected); † marginal ($.05 <$ adjusted $p < .10$).

The most prominent finding was that five of six patterns showing significant or marginally significant differences contained prompting corrective feedback ([t]Cp). These prompting-involved patterns were all significantly more prevalent in HP learners' sessions. The patterns [t]Q → [s]R → [t]Cp and [t]Q → [s]R → [t]Cp → [t]Q were both significantly more frequent in HP sessions (SUP = 28 vs. 15; adjusted p = .041). These sequences represent a pedagogically meaningful cycle: the chatbot poses a question, the student responds (potentially with an error), the chatbot provides prompting feedback encouraging self-correction, and dialogue continues with a follow-up question.

Three additional prompting-related patterns ([s]R → [t]Cp; [s]R → [t]Cp → [t]Q; [s]R → [t]Cp → [t]Q → [s]R) showed marginally significant differences favouring HP sessions (adjusted p = .063), further supporting the association between prompting CF sequences and learning progress.

Besides, some potentially meaningful DA patterns, while marginal or non-significant in the current analysis, may merit further exploration in future studies. For example, [t]Q → [s]D, where students express disagreement following a chatbot question, was marginally more frequent in HP sessions (16 vs. 5 sessions; adjusted $p$ = .079), suggesting greater engagement in meaning negotiation. Although not statistically significant, [t]R → [t]Q → [s]R → [t]A occurred more often in LP sessions, potentially reflecting more surface-level interactional routines that warrant further examination.

## 5     Discussion

### 5.1     RQ1: Potential DA Indicators of Learning Gains

In response to RQ1, overall DA distributions across groups were largely comparable, with two learner DAs emerging as potential indicators differentiating high- and low-gain sessions. Across both groups, interaction was predominantly chatbot-led: the chatbot's contributions were dominated by referential questions ([t]Q; 31.4% of all DA codes) and agreement moves ([t]A; 19.0%), whereas learner contributions were largely responses ([s]R; 27.2%). This distribution suggests that, in the present context, the chatbot largely managed and sustained the interaction, while learners participated mainly in a reactive role.

Within this largely comparable distribution, two learner DAs emerged as potential DA-level indicators of learning gains. High-gain sessions contained significantly more learner-initiated questions ([s]Q; HP: 2.1% vs. LP: 1.0%), indicating greater learner initiative in directing the interaction rather than merely responding. Consistent with prior research, question-asking reflects active engagement in meaning negotiation and allows learners to pursue personally relevant communicative goals [5, 32]. In contrast, low-gain sessions showed significantly higher frequencies of clarification-seeking moves ([s]S; LP: 2.5% vs. HP: 1.3%), suggesting more frequent comprehension difficulties when processing chatbot utterances. Such breakdowns are consistent with reported mismatches between chatbot output complexity and learners' listening comprehension abilities [33] and may constrain productive interaction, as language development depends on learners' ability to process and understand input [34].

From a design perspective, these findings highlight the value of adopting a dialogic lens to identify pedagogically meaningful DAs and support adaptive responses in GenAI-mediated oral practice. For example, learner-initiated questions ([s]Q) can serve as signals of active participation and may be deliberately encouraged, whereas repeated clarification-seeking moves ([s]S) can be used to trigger additional scaffolding, such as rephrasing, slower speech, or brief comprehension checks, to better align chatbot output with learners' comprehension capacities. More broadly, these findings underscore the importance of monitoring learners' DAs during interaction and making in-time adaptations accordingly, as effective learning support depends on context-sensitive decisions informed by ongoing interactional signals [35].

### 5.2  RQ2: Sequential DA Patterns Associated with Learning Gains

In response to RQ2, the most salient finding is that sessions associated with HP learners were characterised by significantly more prompting ([t]Cp) sequences. Two significant patterns in HP sessions, i.e. [t]Q → [s]R → [t]Cp and [t]Q → [s]R → [t]Cp → [t]Q (both p_adj = .041), together with three marginally significant patterns containing [t]Cp, together distinguished HP sessions. In contrast, no significant sequential differences were observed for other CF types, i.e. recasts ([t]Cr) or explicit correction ([t]Ce). These results suggest that prompting, rather than other CF strategies, was more closely associated with higher learning gains in GenAI-mediated oral practice, consistent with previous research showing that prompts are particularly effective in eliciting learner effort and modified output, thereby supporting language development [25, 26].

Importantly, this distinction emerged only at the *sequential* level. No significant group differences were found in the overall frequency of all CF types ([t]Cp, [t]Cr, [t]Ce), indicating that what differentiates HP sessions is not whether CF occurred or how often it occurs, but *how* and *when* it is used within the interactional flow. Closer inspection of the significant patterns further indicated that effective prompting was consistently positioned after learner responses ([s]R), rather than at other interactional points such as following learner-initiated questions ([s]Q). One plausible interpretation is that correcting grammatical errors within learner-initiated questions, instead of engaging with their content, may disrupt interactional flow, consistent with prior work highlighting the importance of feedback timing in dialogue [36].

From a design perspective, chatbot design should specify not only whether particular pedagogical actions are implemented, but also *when and how* they are triggered during interaction [37]. This, in turn, calls for the adoption of a dialogic lens to identify effective (and less effective) DA patterns associated with learning gains, and to use these patterns to inform more evidence-based prompt engineering of chatbots' pedagogical behaviours in GenAI-mediated oral practice.

### 5.3  Limitations, Future Directions, and Design Implications

Several limitations should be acknowledged. First, English oral development is inherently longitudinal and is best understood across multiple sessions rather than single sessions. Accordingly, data were collected over eight weeks, yielding 70 sessions from 12 learners, all of which were manually coded at the turn level to enable a fine-grained analysis of dialogic processes. As sessions from the same learner cannot be assumed to be independent, analyses were aggregated at the learner level, and the intensive effort required for turn-level manual coding inevitably limited the number of learners analysed. The findings should therefore be interpreted as exploratory. Moreover, although human coding is rigorous and yields rich interactional insights, its highly complex and cognitively demanding nature [27] constrains the feasibility of larger-scale analyses.

Future studies should thus include larger learner cohorts to enhance generalisability and explore automated DA classification methods to facilitate such larger-scale analyses.

From a design perspective, despite the exploratory nature of the current study, several meaningful process-oriented design insights can be derived: (1) incorporating comprehension monitoring to detect and respond to clarification requests; (2) adopting dialogue strategies that create more opportunities for learner-initiated questions; and (3) prioritising prompting over explicit correction, with careful specification of when such feedback is delivered during interaction. Overall, these insights highlight the value of adopting a dialogic lens in GenAI voice chatbot research to inform fine-grained chatbot design and development.

## 6      Conclusion

This study examined DA patterns in GenAI-mediated L2 oral practice using both DA-level and sequential analyses. The results indicate that interaction was predominantly chatbot-led, with higher-gain sessions characterised by greater learner-initiated questioning and more effective use of prompting following learner responses, while lower-gain sessions showed more frequent clarification-seeking. These findings suggest that specific student DAs may serve as potential indicators of learning gains, and that learning is mediated not only by which pedagogical moves occur, but also by *how* and *when* they are enacted in interaction. Overall, this highlights the value of a dialogic lens for understanding learning processes underlying GenAI-mediated outcomes and for informing fine-grained design of educational GenAI chatbots.

**Acknowledgments.** We thank our annotators for their dedication throughout the coding process. We are grateful to all participating students and teachers who made this research possible.